\documentclass[conference]{IEEEtran}
\IEEEoverridecommandlockouts    % to override locked commands

\usepackage{multirow}
\usepackage{lipsum}
\usepackage{amsmath}
\usepackage{amssymb}
\usepackage[ruled,vlined]{algorithm2e}
\usepackage{graphicx}
\graphicspath{{./Figures/}}

\usepackage[utf8]{inputenc} % allow utf-8 input
\usepackage[T1]{fontenc}    % use 8-bit T1 fonts
\usepackage{hyperref}       % hyperlinks
\usepackage{url}            % simple URL typesetting
\usepackage{booktabs}       % professional-quality tables
\usepackage{amsfonts}       % blackboard math symbols
\usepackage{nicefrac}       % compact symbols for 1/2, etc.
\usepackage{microtype}      % microtypography
\usepackage{xcolor}         % colors

\usepackage{amsmath,amsfonts,bm}

\newcommand\mypara[1]{\vspace{0.0mm}\noindent\textbf{#1}}

\definecolor{citecolor}{rgb}{0,0.08,0.45}
\definecolor{linkcolor}{RGB}{187,18,26}

\definecolor{graftstereo}{RGB}{238, 75, 63}
\definecolor{raftstereo}{RGB}{255, 160, 31}
\definecolor{egdepth}{RGB}{91, 158, 204}
\definecolor{sdgdepth}{RGB}{111, 192, 112}

\def\eqref#1{equation~\ref{#1}}

\def\1{\bm{1}}

\DeclareMathAlphabet{\mathsfit}{\encodingdefault}{\sfdefault}{m}{sl}
\SetMathAlphabet{\mathsfit}{bold}{\encodingdefault}{\sfdefault}{bx}{n}

\def\sR{{\mathbb{R}}}

\newcommand{\eg}{{\em e.g.}}
\newcommand{\ie}{{\em i.e.}}

\DeclareMathOperator*{\argmax}{arg\,max}

\usepackage{adjustbox}
\usepackage{multirow}
\usepackage{pifont}
\usepackage{wrapfig}
\usepackage[font=footnotesize]{caption}
\usepackage{subcaption}
\let\labelindent\relax
\usepackage{enumitem}
\usepackage[noadjust]{cite}

\usepackage[capitalize]{cleveref}
\crefname{section}{Sec.}{Secs.}
\crefname{table}{Table}{Tables}
\crefname{figure}{Fig.}{Figs.}

\title{\vskip15pt\LARGE \bf Preserving Guidance in Cost-Volume Retrieval\\under Extreme LiDAR Sparsity in Iterative Stereo}

\author{
    \parbox{\textwidth}{%
        \centering
        Jinsu Yoo$^{1}$, Sooyoung Jeon$^{1}$, Zanming Huang$^{1}$,
        Tai-Yu Pan$^{2}$, Wei-Lun Chao$^{1,3}$%
    }%
    \thanks{$^{1}$The Ohio State University, Columbus, OH, USA.
    \texttt{\{yoo.503, jeon.193, huang.5758, pan.667\}@osu.edu}}
    \thanks{$^{2}$Google, Sunnyvale, CA, USA.}
    \thanks{$^{3}$Boston University, Boston, MA, USA. \texttt{chao209@bu.edu}}
}

\begin{document}
	
\maketitle
\thispagestyle{empty}
\pagestyle{empty}
	
\begin{abstract}
While accurate LiDAR depth has been shown to improve stereo matching, high-end LiDAR remains costly and difficult to deploy at scale, motivating guidance from sparse LiDAR measurements.
In this paper, we revisit how extremely sparse LiDAR can guide iterative stereo by examining its cost-volume retrieval mechanism. 
Our analysis shows that LiDAR guidance degrades sharply when only a few hundred points are available, and we provide a signal-processing explanation for why iterative stereo fails to fully exploit such sparse input. 
This insight leads to a simple yet highly effective remedy: pre-filling the initial disparity map so that cost-volume retrieval becomes more reliable under extremely sparse LiDAR. 
We find that pre-filling is also effective when injecting LiDAR depth into image features via early fusion, but for a fundamentally different reason, necessitating a distinct pre-filling approach.
Combining both forms of guidance, our proposed \textbf{G}uided \textbf{RAFT-Stereo} (\textbf{GRAFT-Stereo}) achieves substantial improvements over existing LiDAR-guided stereo methods across multiple datasets. Project page: \url{https://jinsuyoo.info/graft-stereo}.
\end{abstract}

\section{Introduction}
\label{sec:intro}

\begin{figure}[t]
  \centering
  \begin{subfigure}[t]{\linewidth}
    \centering
    \includegraphics[width=0.5\linewidth]{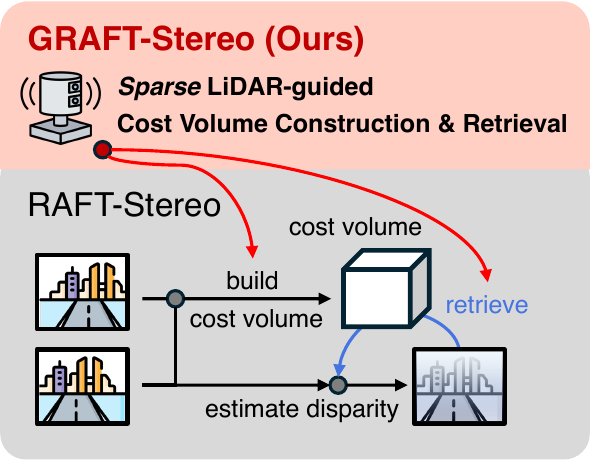}
    \caption{\footnotesize Method overview.}
    \label{fig:1a}
  \end{subfigure}
  \vskip 3pt
  \begin{subfigure}[t]{\linewidth}
    \centering
    \includegraphics[width=\linewidth]{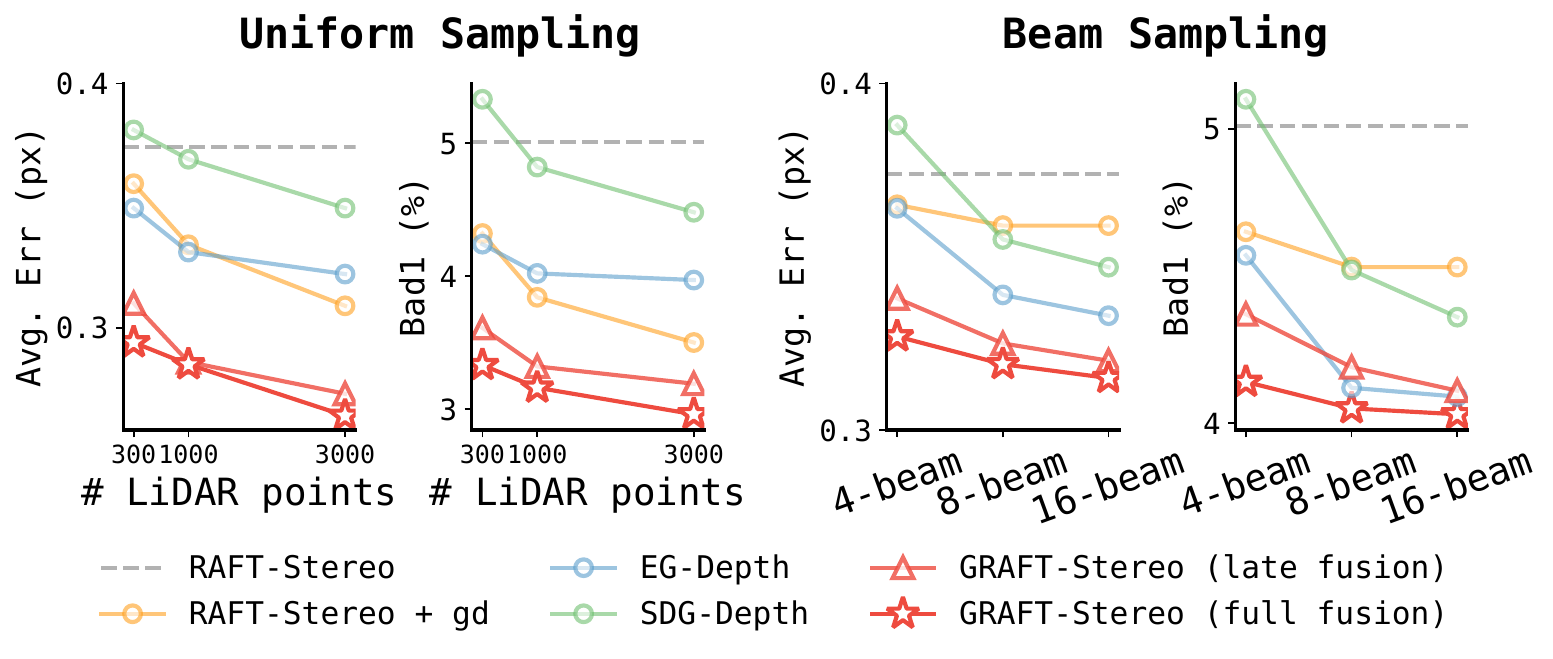}
    \vskip-5pt
    \caption{\footnotesize Performance comparison under various LiDAR sparsity conditions.}
    \label{fig:1b}
  \end{subfigure}
  \caption{\textbf{Overview of GRAFT-Stereo.}
  \textbf{(a):} {\color{graftstereo}GRAFT-Stereo} integrates very sparse LiDAR points with minimal modifications to {\color{raftstereo}RAFT-Stereo}~\cite{lipson2021raftstereo}.
  \textbf{(b):} {\color{graftstereo}GRAFT-Stereo} achieves a substantial reduction in disparity error on the KITTI benchmark \cite{uhrig2017kittidc} across various LiDAR sparsity conditions, surpassing the SotA LiDAR-guided stereo methods, {\color{egdepth}EG-Depth}~\cite{xu2023egdepth} and {\color{sdgdepth}SDG-Depth}~\cite{li2024sdgdepth}.
  gd: naive LiDAR-guidance; Avg. Err: average disparity error over all pixels; Bad1: percentage of pixels with a disparity error greater than 1.0 pixel.}
  \vskip-10pt
  \label{fig:teaser}
\end{figure}

Reliable dense depth is a core sensing primitive for outdoor robotics and autonomous driving, supporting tasks such as obstacle avoidance, mapping, and long-range perception. Stereo cameras are attractive due to their low cost and wide coverage, motivating continued interest in high-quality stereo matching. 

In recent years, RAFT-Stereo variants (\cite{lipson2021raftstereo, xu2023igev, wang2024selective, li2022crestereo}), \ie, iterative refinement-based methods, have demonstrated excellent performance in estimating pixel disparity from rectified stereo image pairs. 
Unlike CNN-based~(\cite{chang2018psmnet, kendall2017gcnet, yang2019hierarchical}) and Transformer-based methods~(\cite{li2021sttr, guo2022context}), RAFT-Stereo offers a distinctive advantage: \emph{it supports anytime prediction}, letting users trade inference time for accuracy through the number of refinement iterations.
However, like all stereo methods, RAFT-Stereo struggles in outdoor scenes with complex content and distant ranges, where closing the gap between estimated and actual depth remains an open challenge.

LiDAR-guided stereo offers a promising solution to this challenge by \emph{leveraging sparse but highly accurate LiDAR depth to improve dense stereo depth prediction.} Previous work has primarily relied on convolutional neural networks~(\cite{choe2021volumetric, xu2023egdepth, li2024sdgdepth,li2025sparse}). In this paper, we explore how LiDAR input can be integrated into the RAFT-Stereo framework. \emph{Our goal is to develop a principled approach that introduces minimal changes to RAFT-Stereo, preserving its inherent strengths.}

We first investigate where to inject LiDAR depth. 
We identify the \emph{initial disparity map} in RAFT-Stereo as an ideal candidate, as it processes disparity values rather than features and is iteratively refined. This led us to hypothesize that LiDAR depth, once converted to disparity, could be seamlessly injected into the map without additional encoding~\cite{zhao2022sparse}.
Our initial attempt, while vanilla, already yielded promising results. 
Under the standard stereo-LiDAR fusion setting with 64-beam LiDAR points~(\cite{xu2023egdepth, li2024sdgdepth}), replacing the default all-zero initial disparity with LiDAR-derived values effectively reduces the average disparity error to $0.317$ for RAFT-Stereo, compared to $0.300$ for the state-of-the-art (SotA) LiDAR-guided method~\cite{li2024sdgdepth}. 

However, assuming access to a 64-beam LiDAR is often impractical, as such sensors cost over ten thousand dollars~\cite{geiger2012kitti}, contradicting a core advantage of stereo depth: affordability. Ideally, we seek to leverage LiDAR input with minimal added cost, recognizing that only a limited number of points may be available in practice. Unfortunately, as shown in~\cref{fig:1b} and~\cref{fig:num-points}, \emph{the disparity error of} \emph{\color{raftstereo}{vanilla LiDAR-guided RAFT-Stereo}} \emph{increases sharply as the LiDAR becomes very sparse.}

To better understand this limitation, we conduct a detailed analysis and find that the performance drop is not solely due to the reduced point count but also to the \emph{internal mechanics of RAFT-Stereo}, which unintentionally suppresses the impact of LiDAR guidance. Specifically, RAFT-Stereo uses the (initial/previous) disparity map to index cost volume features during refinement. In extremely sparse settings, most pixels are initialized with zero disparity, while only a few receive accurate values from LiDAR. As a result, the features retrieved from these accurate disparities appear as high-frequency outliers relative to the dominant zero-disparity initialization and are subsequently attenuated by the recurrent convolution operations.

Building on this insight, we propose a simple yet highly effective solution: \emph{pre-filling the LiDAR-missing pixels} in the initial disparity map. Remarkably, by applying a purely image-processing-based interpolation method~\cite{ku2018ipfill} to densify extremely sparse disparities (\eg, $300$ points), the resulting LiDAR-guided RAFT-Stereo surpasses its counterpart that uses 64-beam LiDAR without pre-filling.

We further investigate alternative locations within RAFT-Stereo for incorporating LiDAR information. Notably, appending the 3D coordinates of each LiDAR point to the RGB values of the corresponding pixels proves effective, as it explicitly reinforces stereo correspondence.
For this form of LiDAR guidance, pre-filling LiDAR-missing pixels remains beneficial. However, it demands a more advanced approach to provide more accurate depth estimates than simple interpolation.

By integrating both forms of LiDAR guidance with their respective pre-filling strategies, our proposed \textbf{G}uided \textbf{RAFT-Stereo} (\textbf{\color{graftstereo}{GRAFT-Stereo}}) achieves SotA stereo matching performance under extremely sparse LiDAR conditions (see~\cref{fig:teaser}). Extensive experiments on three outdoor datasets---KITTI Depth Completion~\cite{uhrig2017kittidc}, VKITTI2~\cite{cabon2020vkitti}, and MS2~\cite{shin2023ms2}---consistently validate its effectiveness. 

To summarize, our main contributions are as follows:
\begin{itemize}[nosep, topsep=2pt, parsep=2pt, partopsep=2pt, leftmargin=*]

\item We analyze RAFT-Stereo as a LiDAR-guided stereo framework and identify why its naïve use of \emph{extremely sparse} LiDAR yields limited gains, linking this degradation to its cost-volume retrieval mechanism.

\item We introduce \textbf{GRAFT-Stereo}, which enables RAFT-Stereo to effectively leverage extremely sparse LiDAR through a depth pre-filling strategy, without introducing any sophisticated architectural components.

\item We evaluate \textbf{GRAFT-Stereo} on three outdoor datasets---KITTI Depth Completion, VKITTI2, and MS2---achieving consistent improvements over existing LiDAR-guided stereo methods in the extremely sparse LiDAR setting.

\end{itemize}
\section{Related Work}
\label{sec:related-work}

\mypara{Stereo matching.}
Traditional stereo matching methods focused on handcrafted features and prior modeling~(\cite{zabih1994non,yang2010constant,veksler2005stereo,yang2008stereo,hirschmuller2007stereo}), whereas deep learning has since transformed the field~(\cite{tosi2023nerf,chang2023domain,weinzaepfel2023crocov2,liu2022graftnet,jing2023uncertainty,tian2023dps,poggi2024federated,xu2023accurate,bartolomei2023active,guan2024neural,guo2023openstereo}). 
Recent methods can be broadly categorized into three groups: CNN-based, Transformer-based, and iterative approaches.
CNN-based methods~(\cite{chang2018psmnet,kendall2017gcnet,yang2018segstereo,yang2019hierarchical}) construct and aggregate a cost volume, typically using 3D convolutions---an approach achieving high accuracy but remains computationally intensive.
Transformer-based methods~(\cite{li2021sttr,guo2022context}), inspired by the success of Vision Transformers~\cite{dosovitskiy2020image}, replace conventional cost volume processing with optimal transport and attention mechanisms.
Iterative methods~(\cite{lipson2021raftstereo,xu2023igev,li2022crestereo,zhao2023high,wang2024selective,zhao2022eai,zeng2023parameterized,wen2025foundationstereo}), drawing from RAFT~\cite{teed2020raft}, process cost volumes through iterative feature retrieval, offering higher efficiency and enabling anytime prediction.
While recent foundation stereo models~\cite{wen2025foundationstereo} pursue universal representations via large-scale pretraining, they are \emph{orthogonal} to our goal: diagnosing and improving RAFT-Stereo's behavior under extremely sparse LiDAR. 

\mypara{LiDAR-guided stereo.}
To overcome the limitations of standard stereo matching, guided stereo incorporates additional reliable depth measurements, typically from a LiDAR sensor~(\cite{cheng2019noise,mai2021sparse,choe2021volumetric,wang20193d,zhang2022slfnet,pilzer2023expansion,bartolomei2023active}). 
Prior work has explored various integration strategies, including cost volume adjustment with sparse depth~(\cite{poggi2019guided,huang2021s3}) and fusion at the input or feature level~(\cite{park2018high,zhang2020listereo,xu2023egdepth,li2024sdgdepth,li2025sparse}). 
In this work, we study how sparse LiDAR can guide RAFT-Stereo~\cite{lipson2021raftstereo}, leveraging its modular design and iterative refinement capability to target cost-efficient deployment under extremely sparse input (\eg, 300 points). 
While pre-filling is often used to densify sparse depth, our use of pre-filling serves a different purpose: stabilizing RAFT-Stereo’s cost-volume retrieval under extremely sparse LiDAR rather than improving depth quality itself.

\mypara{Depth completion.}
Our work is inspired by depth completion, which estimates dense depth from \emph{monocular} images and sparse LiDAR input~(\cite{shivakumar2019dfusenet,qiu2019deeplidar}).
Prior efforts have primarily focused on leveraging sparse depth through feature extraction~(\cite{hu2021penet,tang2020learning,liu2023mff}) or spatial propagation~(\cite{park2020non,liu2017learning,cheng2018depth,wang2023lrru}), with pre-filling strategies also being explored~\cite{liu2021fcfr}.
We systematically analyze RAFT-Stereo's \emph{cost volume retrieval mechanism}, identify its core limitation in utilizing LiDAR guidance, and propose pre-filling as a simple yet highly effective solution.
\section{Background: Stereo Matching for Disparity Estimation 
 and RAFT-Stereo}
\label{sec:background}

Given a pair of rectified images $I_l$ and $I_r$ from parallel cameras, \textbf{stereo matching} estimates the horizontal pixel disparity by identifying corresponding pixels across the two images. Let $D(h, w)$ denote the disparity at pixel $(h, w)$ in the left image $I_l$, where $h$ and $w$ represent the vertical and horizontal coordinates, respectively. Ideally, $I_l(h, w)$ and $I_r\left(h, w - D(h, w)\right)$ correspond to the same 3D point in the scene. The disparity map $D$ can be converted into a depth map $Z$ using the known focal length $f$ and camera baseline $b$ (\ie, the distance between the cameras), as
$Z(h, w) = \frac{f \cdot b}{D(h, w)}$.

Traditional methods typically construct a 3D cost volume $C$ that encodes pixel-wise correspondences along horizontal scanlines, where $C(h,w,w')$ represents the similarity between pixels $I_l(h,w)$ and $I_r(h,w')$. Given a reliable cost volume, the disparity at pixel $(h, w)$ can be estimated as $D(h, w) = w - \argmax_{w'} C(h,w,w')$. However, the cost volume is often degraded by challenging conditions such as lighting variations, occlusions, or textureless regions. To mitigate these issues, standard approaches apply computationally intensive 3D convolutions to denoise the cost volume, often incorporating smoothness priors to regularize the final disparity estimates.
 
\textbf{RAFT-Stereo}~\cite{lipson2021raftstereo} bypasses 3D convolutions by iteratively refining disparity estimates. At each step, it extracts a local region of the 3D cost volume $C$ centered on the current estimate $D$, and refines the disparity using recurrent 2D convolutional networks. This approach regularizes the cost volume selectively---only where needed---yielding both accuracy and efficiency. 
Let $H$ and $W$ denote the height and width of the image, respectively. Conceptually, a local region $S\in\sR^{H \times W \times (2K{+}1)}$ can be constructed by retrieving the $2K{+}1$ cost values centered at the current estimate, \ie, at $w' = w - D(h, w)$ for each pixel $(h, w)$:
\begin{align}
S(h, w, k) = C(h, w, w - D(h, w) + k),  k \in [-K, K]. \label{eq:retrieval}
\end{align}
The resulting $S$ can be processed by 2D convolutions, treating the 3\textsuperscript{rd} dimension as the channel axis.

Without loss of generality, a RAFT-Stereo model can be implemented as follows. Given $I_l$ and $I_r$, it 
first extracts feature maps $X_l$ and $X_r$ using a shared 2D CNN backbone (\eg, ResNet~\cite{he2016deep}).
Each feature map is of size $H \times W \times J$, where $J$ is the feature dimensionality. 
Next, a 3D correlation volume is constructed by:
\begin{align}
C(h, w, w') = \sum_j X_l(h, w, j) \cdot X_r(h, w', j). \label{eq:correlation}
\end{align}
Then, starting from an initial disparity map $D^{(0)}$ (typically set to zero), the model iteratively refines the estimate.
At iteration $t$, it retrieves $S^{(t)}$ from $C$ based on the current estimate $D^{(t-1)}$, and applies a recurrent 2D CNN to predict a residual disparity $\Delta^{(t)}$. The estimate is then updated as:
\begin{align}
D^{(t)} = D^{(t-1)} + \Delta^{(t)}. \nonumber
\end{align}
During training, the model is optimized to minimize the difference between $D^{(t)}$ and the ground-truth disparity map $D^\star$, for all $t\in [1, T]$, where $T$ is the maximum number of iterations. At test time, the model supports anytime prediction: each intermediate output $D^{(t)}$ serves as a valid estimate, with higher $t$ generally yielding greater accuracy.

\section{Setup, Goal, and Preliminary Results}
\label{sec:setting}

This paper investigates how to leverage sparse yet accurate LiDAR depth to enhance dense stereo matching. Our study follows a systematic, step-by-step analysis. We begin by outlining the setup and goals.

\begin{figure}[t]
  \centering
  \includegraphics[width=.5\linewidth]{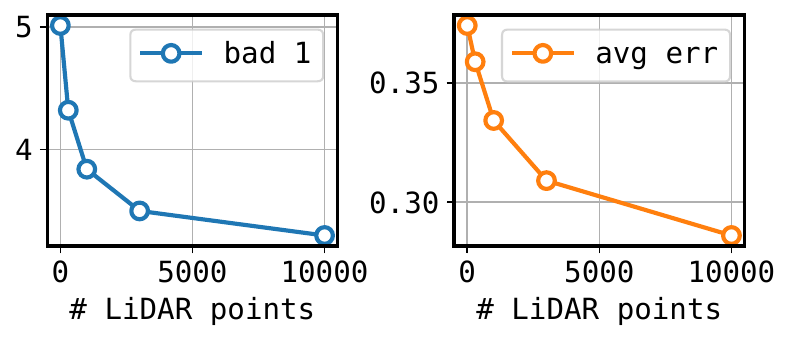}
  \vskip-5pt
  \caption{\textbf{Performance of RAFT‑Stereo~\cite{lipson2021raftstereo} with different numbers of LiDAR points.}
  The disparity error increases drastically as guidance becomes sparse.}
  \vskip-10pt
  \label{fig:num-points}
\end{figure}

\subsection{Experiment setup}

\mypara{Datasets.}
We focus on stereo-LiDAR fusion in outdoor settings and primarily evaluate on the KITTI Depth Completion dataset~\cite{uhrig2017kittidc}, which consists of 42,949 training and 3,426 validation stereo image pairs with corresponding 64-beam LiDAR scans from driving scenes. Each image has a resolution of 375$\times$1242. 
Ground-truth depth is semi-densified by aggregating LiDAR points from neighboring frames, yielding approximately 100k valid points per frame. Following previous works~(\cite{xu2023egdepth, li2024sdgdepth}), we report results on the validation set. We further demonstrate the effectiveness of our method on the VKITTI2~\cite{cabon2020vkitti} and MS2~\cite{shin2023ms2} datasets.

\mypara{Evaluation metrics.}
To evaluate disparity estimates, we compute error rates: the percentage of pixels with a disparity error greater than one (\ie, Bad1) or two (\ie, Bad2) pixels from the ground-truth disparity map. We also report the average disparity error over all pixels.
The maximum disparity is set to 192.
For depth evaluation, we report both the mean absolute error (MAE) and the root mean squared error (RMSE) from the ground-truth semi-dense depth map in millimeters (mm).

\mypara{Implementations.}
We generally follow the official implementation of RAFT-Stereo~\cite{lipson2021raftstereo}, including objective functions and hyperparameters. We evaluate on $t=32$ during inference, unless stated otherwise. The model is trained for 100k iterations using a learning rate of  \(2 \times 10^{-3}\) with one-cycle decay. We set the batch size per GPU to 4 and use two NVIDIA A100 GPUs for all experiments.
For the analysis in~\cref{ss:late} of the main paper, we randomly crop the input images to a size of $336\times1120$. In~\cref{ss:early}, we reduce the crop size to $224\times784$ due to GPU memory constraints.

To synthesize low-beam LiDAR, we follow the implementation provided by~\cite{zhao2021surface} and sample more LiDAR lines from the lower part of the scene, similar to the strategy used in~\cite{you2019pseudo}.

\subsection{Goal: affordable settings with limited points}

We aim to investigate LiDAR-guided RAFT-Stereo with minimal additional cost. As discussed in~\cref{sec:intro}, a 64-beam LiDAR sensor---capable of producing over 10,000 points per frame---is prohibitively expensive for many applications. Thus, we focus on leveraging much sparser LiDAR guidance that simulates the output of lower-cost sensors. We consider two settings:
(1) uniformly subsampling the ground-truth semi-dense depth map, following~\cite{huang2021s3}; and (2) synthesizing a reduced-beam LiDAR response from the original 64-beam data, inspired by~\cite{you2019pseudo}. In extreme cases, a single LiDAR scan may contain only a few hundred points---orders of magnitude fewer than the number of pixels per image (\eg, over 400k in the KITTI dataset).

\subsection{Preliminary results}

\begin{table}[t]
  \centering
  \caption{\textbf{Vanilla LiDAR guidance for RAFT‑based stereo methods with 64‑beam sensors.}
  Iterative refinement‑based methods show great potential for LiDAR guidance even
  \textit{without architectural changes}. SDG‑Depth~\cite{li2024sdgdepth} is the
  state‑of‑the‑art LiDAR‑guided stereo method.}
  \label{tab:64-beam}
  \renewcommand{\arraystretch}{1.0}
  \begin{adjustbox}{width=.9\linewidth}
    \begin{tabular}{lcccc}
      \toprule
      \textbf{Method} & \textbf{Guidance} &
        \textbf{Bad1 (\%)} & \textbf{Bad2 (\%)} & \textbf{Avg. Err. (px)} \\
      \midrule
      RAFT‑Stereo~\cite{lipson2021raftstereo}          & \ding{55} & 5.01 & 0.41 & 0.374 \\
      RAFT‑Stereo~\cite{lipson2021raftstereo}          & 64‑beam    & 4.12 & 0.37 & 0.317 \\
      Selective‑RAFT~\cite{wang2024selective}          & \ding{55} & 5.00 & 0.39 & 0.379 \\
      Selective‑RAFT~\cite{wang2024selective}          & 64‑beam    & 4.05 & 0.36 & 0.318 \\ \midrule
      \textcolor{gray}{SDG‑Depth~\cite{li2024sdgdepth}} & \textcolor{gray}{64‑beam} &
        \textcolor{gray}{3.60} & \textcolor{gray}{0.36} & \textcolor{gray}{0.300} \\
      \bottomrule
    \end{tabular}
  \end{adjustbox}
  \vskip-5pt
\end{table}

\mypara{Where to inject LiDAR points?}  
The modular design of RAFT-Stereo allows LiDAR information to be integrated at different stages. In addition to early fusion in image space and intermediate fusion in feature space, \emph{RAFT-Stereo offers a unique late fusion opportunity via its initial disparity map.} This map is typically initialized to zero, but its iterative refinement process suggests that more accurate initialization could improve performance. This insight is supported by recent work in optical flow~\cite{wang2024sea}, which shows that better initialization enhances RAFT-based results. Motivated by this, we explore initializing RAFT-Stereo's disparity map using sparse LiDAR depth.

\mypara{Results with 64-beam LiDAR.} 
We begin with the standard dense LiDAR-guided stereo setting. 
We inject LiDAR-derived disparities into the corresponding pixel locations of the initial map, and re-train RAFT-Stereo \cite{lipson2021raftstereo} and the recently proposed variant Selective-RAFT~\cite{wang2024selective}.
Surprisingly, {this simple strategy already yields performance comparable to the SotA LiDAR-guided stereo method~\cite{li2024sdgdepth}, underscoring the strong compatibility of RAFT-Stereo with LiDAR guidance (\cref{tab:64-beam}).}

\mypara{Results with fewer LiDAR points.}
We next investigate the impact of injecting fewer points into the initial map. Separate models are trained under varying LiDAR sparsity. As shown in~\cref{fig:num-points}, the performance gains from LiDAR guidance drop sharply as the number of points decreases, highlighting a key challenge for deploying LiDAR-guided RAFT-Stereo in practical, low-cost settings.

\begin{figure}[t]
\centering
\includegraphics[width=\linewidth]{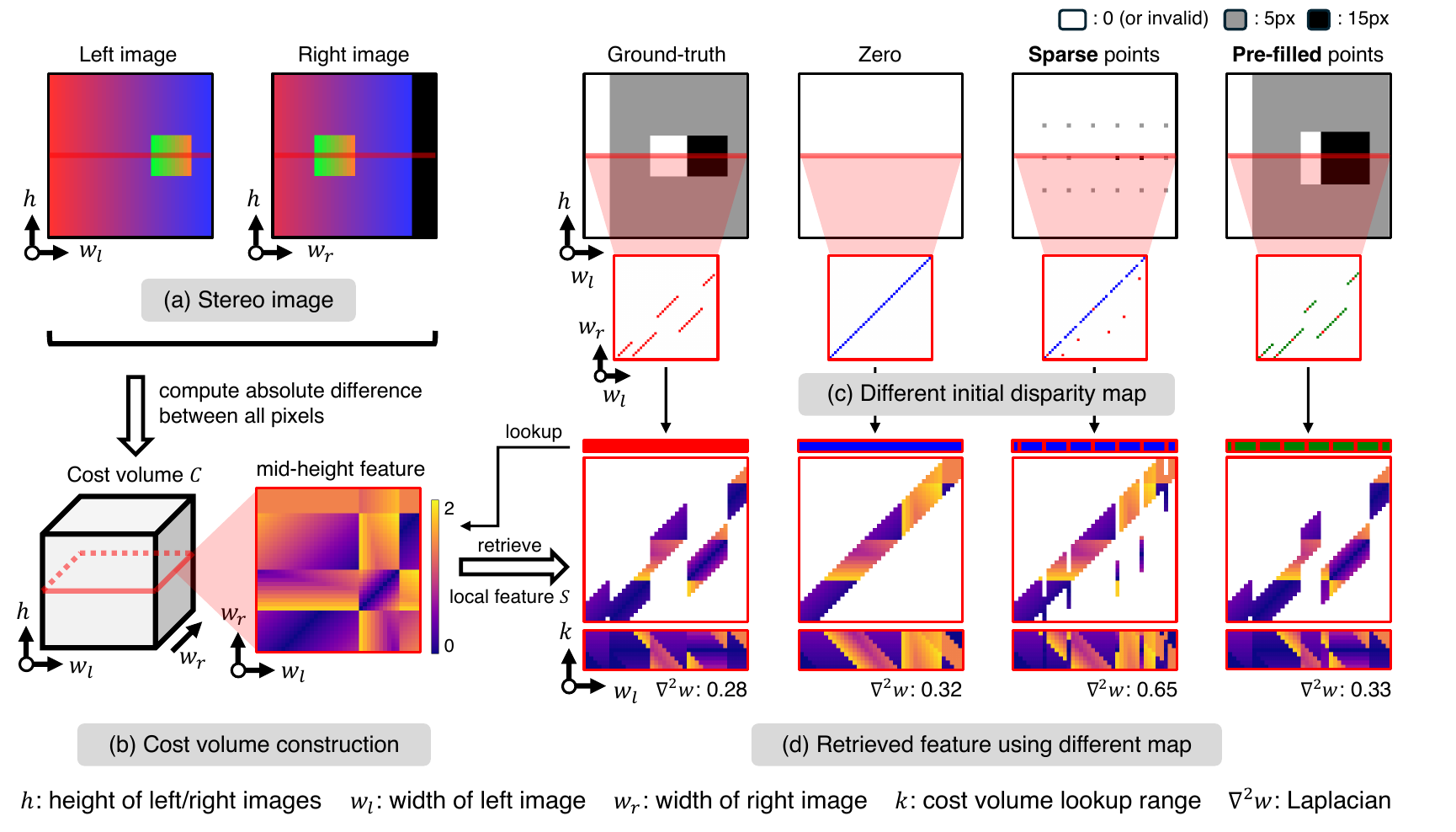}
\vskip-3pt
\caption{\textbf{Illustration of cost volume retrieval in RAFT-Stereo~\cite{lipson2021raftstereo}.} We compare four different initial disparity maps for feature retrieval: ground-truth disparity, default zero initialization, sparse ground truth \emph{mimicking sparse LiDAR}, and a nearest-neighbor-filled version of the sparse map, as shown in (c). In (d), the retrieved features using the sparse ground truth exhibit unintended horizontal artifacts due to abrupt disparity changes, which limit the effectiveness of informed initialization. In contrast, pre-filling the missing disparities helps reinforce spatial consistency and better preserve useful guidance within the retrieved features.}
\label{fig:retrieval}
\vskip-8pt
\end{figure}

\begin{figure}[t]
\centering
\includegraphics[width=\linewidth]{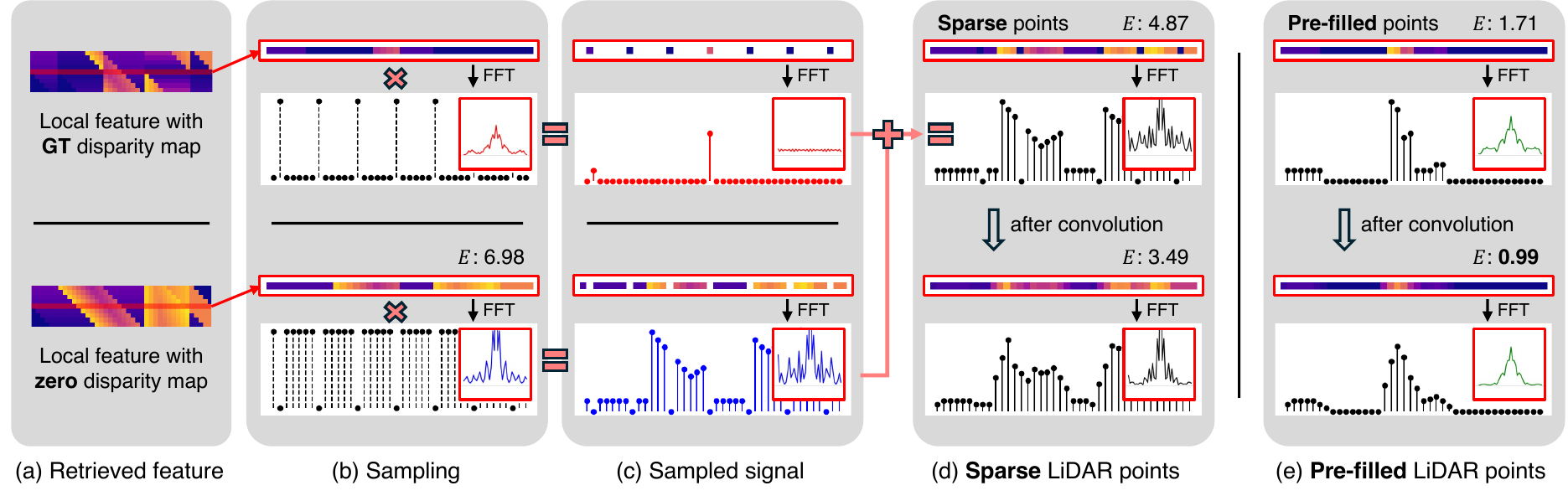}
\vskip-3pt
\caption{\textbf{Analysis of cost volume features retrieved with sparse guidance.} \textbf{(a)-(d):} The retrieved features using sparse disparity guidance (as in (d), top) can be viewed as a sampled combination of those obtained using ground-truth and zero-initialized maps (\ie, (a)-(c)). When the guidance is too sparse, the retrieved features are dominated by the zero-initialization, both before and after applying a low-pass filter---essentially burying useful information from sparse LiDAR points. \textbf{(e):} Pre-filling the missing disparity values enables effective guidance propagation. FFT: Fast Fourier Transform; $E$: L2 distance to features retrieved with the ground-truth map.}
\vskip -5pt
\label{fig:spectrum}
\end{figure}
\section{Sparse LiDAR-Guided RAFT-Stereo}

The aforementioned performance degradation motivates us to analyze RAFT-Stereo in depth, aiming to recover the substantial gains enabled by dense LiDAR guidance.

\begin{figure}[t]
\centering
\includegraphics[width=\linewidth]{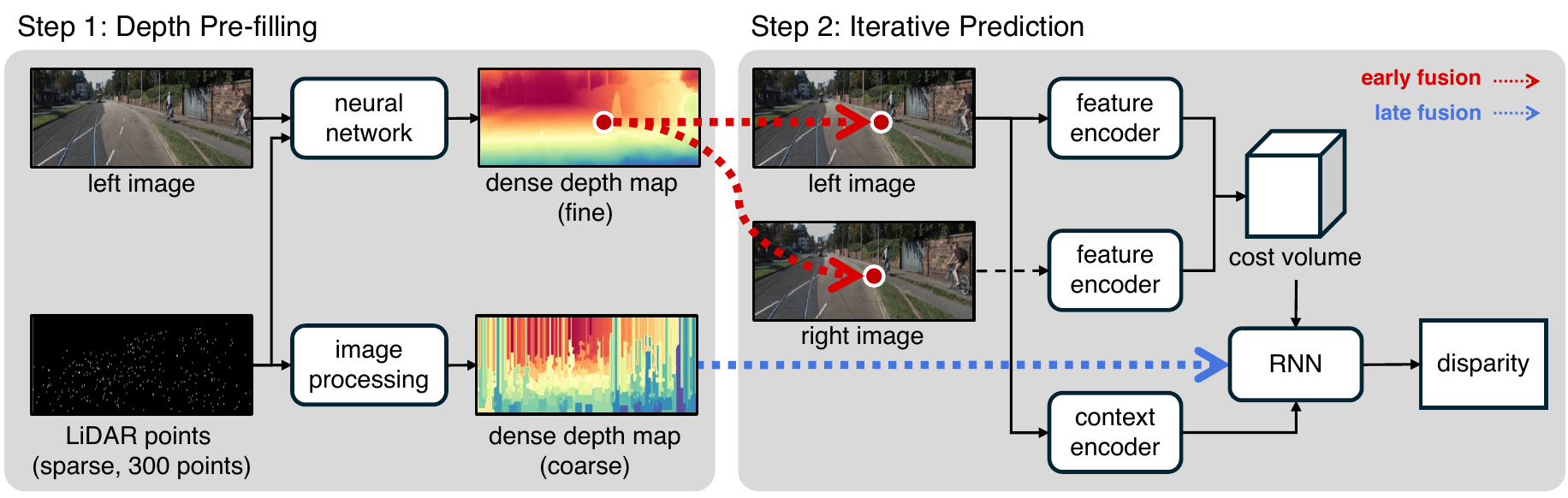}
% \vskip -5pt
\caption{\textbf{Illustration of GRAFT-Stereo.} Given extremely sparse LiDAR points (\eg, 300 points), we systematically identify two complementary strategies to incorporate the LiDAR information into RAFT-Stereo: early fusion and late fusion. Our analysis reveals that \emph{depth pre-filling} is the key to leveraging extremely sparse LiDAR guidance for effective stereo matching.}
\label{fig:overall}
\vskip -10pt
\end{figure}

\subsection{Sparse guidance in the initial disparity map}

We first analyze how LiDAR-derived disparities in the initial disparity map influence RAFT-Stereo. As shown in \cref{eq:retrieval}, RAFT-Stereo uses previous disparity estimates as indices to retrieve \emph{local} cost volume features for refining the estimates. This process can be interpreted as an \emph{iterative local search} for the true disparity value at each pixel, making its performance likely sensitive to the initialization.

In the absence of prior information, RAFT-Stereo typically initializes with a zero disparity map, resembling an uninformed search. In contrast, LiDAR-derived disparities provide guidance for a more informed initialization, akin to an informed search.

The challenge, however, arises as LiDAR measurements become increasingly sparse. In such cases, naively injecting LiDAR-derived disparity values only into the corresponding pixels leaves large portions of the map still initialized with zeros, limiting the benefit of informed initialization. Upon closer inspection of RAFT-Stereo's architecture, we observe that its internal mechanisms may inadvertently suppress or ignore LiDAR guidance when input is extremely sparse.

\mypara{Toy example.}
To illustrate this effect, we develop a toy example using a pair of synthesized images shown in~\cref{fig:retrieval}a. The scene comprises a background and a foreground object, each exhibiting gradually changing colors---red to blue for the background and green to orange for the foreground.
The true disparity values are 15 for the foreground and 5 for the background, as shown in~\cref{fig:retrieval}c.
 
We first construct RAFT-Stereo's cost volume, as shown in~\cref{fig:retrieval}b, using the \textbf{L1 distance} between RGB pixel values instead of correlation to improve visualization clarity. We then retrieve cost volume features using different initial disparity maps, shown in~\cref{fig:retrieval}c, resulting in different local regions $S$.
Specifically, we consider four initializations: the full ground-truth map, a zero-initialized map, a sparse ground-truth map \emph{mimicking sparse LiDAR}, and its nearest-neighbor-filled version. For visualization, we slice the 3D cost volume and retrieved regions at the vertical midpoint (\cref{fig:retrieval}b right, and bottom of \cref{fig:retrieval}d).

\mypara{Sparse LiDAR points inject {unintentional noise}.}
While ground-truth disparities guide the retrieval of low-cost features (darker colors), sparsity introduces abrupt changes in the retrieved features, resulting in horizontal discontinuities. We quantify this spatial irregularity using the Laplacian $\nabla^{2}w$, observing a much higher value for the sparse ground-truth than the full map (0.65 \textit{vs.} 0.28).

We hypothesize that such spatial discontinuities act as noise, degrading RAFT-Stereo's performance by making subsequent 2D convolutions for denoising and refinement more difficult. To test this, we conduct an experiment on KITTI, comparing a zero-initialized disparity map with and without added noise at every other pixel. As shown in~\cref{tab:map-init}, spatial noise in the initial disparity map---and thus in the retrieved features---leads to a measurable drop in RAFT-Stereo's accuracy.

\begin{table}[t]
  \centering
\caption{\textbf{Spatial discontinuities in cost-volume retrieval degrade RAFT-Stereo.} 
We compare with and without noise in zero-initialized disparities.}
  \label{tab:map-init}
  \renewcommand{\arraystretch}{1.0}
  \begin{adjustbox}{width=.65\linewidth}
    \begin{tabular}{lccc}
      \toprule
      \textbf{Method} & \textbf{Noise} &
        \textbf{Bad1 (\%)} & \textbf{Avg. Err. (px)} \\ \midrule
      RAFT‑Stereo~\cite{lipson2021raftstereo} & \ding{55} & 5.01 & 0.374 \\
      RAFT‑Stereo~\cite{lipson2021raftstereo} & \ding{51} & 5.36 & 0.387 \\ \bottomrule
    \end{tabular}
  \end{adjustbox}
\end{table}

\begin{table}[t]
  \centering
  \vskip-5pt
\caption{\textbf{Performance of the two depth pre-fill methods alone,} before applying RAFT-Stereo on the pre-filled disparity maps.}
  \label{tab:prefill}
  \renewcommand{\arraystretch}{1.0}
  \begin{adjustbox}{width=.65\linewidth}
    \begin{tabular}{lccc}
      \toprule
      \textbf{Method} & \textbf{Bad1~(\%)} & \textbf{Bad2~(\%)} &
        \textbf{Avg. Err. (px)} \\ \midrule
      IP‑Basic~\cite{ku2018ipfill} & 52.51 & 37.71 & 2.772 \\
      Neural net                    & 17.59 &  4.22 & 0.700 \\ \bottomrule
    \end{tabular}
  \end{adjustbox}
  \vskip-5pt
\end{table}

\mypara{RAFT-Stereo struggles to \textit{propagate} sparse guidance.}
One intuitive solution to this challenge is to propagate features retrieved via LiDAR guidance to nearby pixels, thereby reducing spatial discontinuities. At first glance, this could be achieved if RAFT-Stereo learns a low-pass filter to interpolate or diffuse the retrieved features. However, unlike conventional interpolation, the features at LiDAR-missing pixels are not zero-valued. Instead, even zero disparities retrieve features from the cost volume, as is standard in RAFT-Stereo. As a result, when LiDAR points are too sparse, their contributions are overwhelmed by features associated with zero disparities, causing the guided features to resemble high-frequency noise. Consequently, applying a low-pass filter fails to propagate the sparse guidance and instead attenuates it.
 
\cref{fig:spectrum} illustrates this effect using the toy example, where the quality of the retrieved features is measured by their L2 distance to those obtained using the full ground-truth disparity map: a smaller distance indicates higher-quality retrieval. Panels (a)-(c) decompose the features retrieved under sparse guidance into components originating from the ground-truth and zero-disparity maps, along with their frequency responses. Panel (d) shows that, both before and after applying a low-pass filter, the retrieved features remain dominated by the zero-disparity component, which inadvertently suppresses the informative contributions from the sparse ground-truth/LiDAR points.

\begin{figure}[t]
\centering
% \vskip -15pt
\includegraphics[width=.8\linewidth]{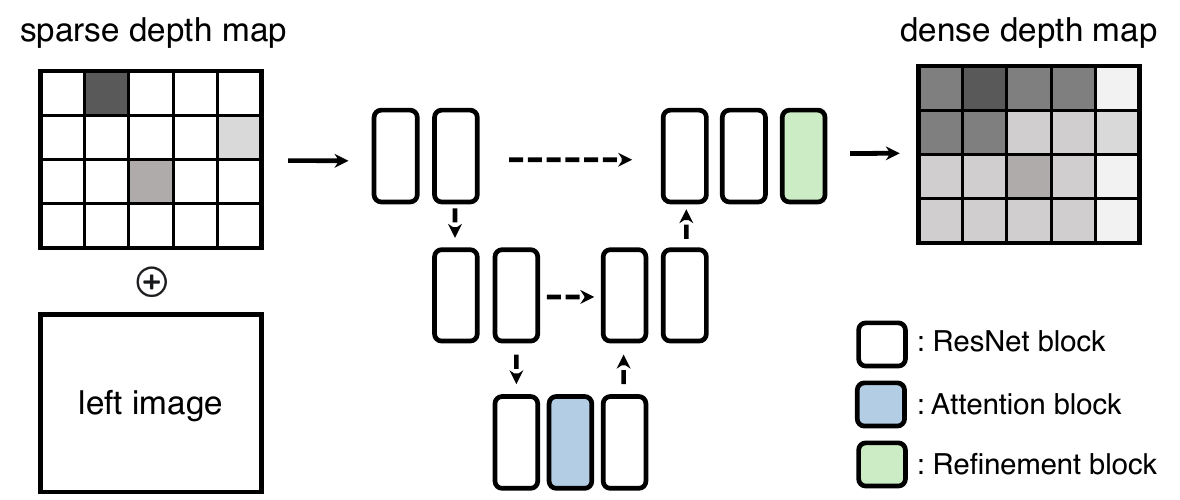}
\caption{\textbf{The neural network architecture for depth pre-filling,} using non-local spatial propagation~\cite{park2020non} for the refinement block.}
\label{fig:unet}
\end{figure}

\subsection{Depth pre-filling improves feature retrieval}
\label{ss:late}
To effectively leverage sparse LiDAR guidance without modifying the core architecture of RAFT-Stereo, we propose a \emph{pre-densification step} to fill in missing disparity values.
Namely, rather than relying on RAFT-Stereo to handle discontinuous features from sparse initial disparity maps, we smooth the disparity map. This reduces horizontal discontinuities in the retrieved features while preserving LiDAR guidance.
As shown in~\cref{fig:retrieval}d, we apply nearest-neighbor interpolation to pre-fill LiDAR-missing pixels in the initial disparity map. This significantly reduces spatial discontinuity in the retrieved local region, lowering the Laplacian value $\nabla^{2}w$ from 0.65 to 0.33. The effectiveness of this strategy is further illustrated in~\cref{fig:spectrum}e, where the L2 distance of the retrieved features decreases sharply, indicating that the zero-disparity component no longer dominates feature retrieval.

\begin{table}[t]
% \vskip -15pt
\centering
\caption{\textbf{Pre-filling the initial disparity map notably improves LiDAR-guided RAFT-Stereo.} We show the results by both the image processing-based~\cite{ku2018ipfill} and neural network-based pre-filling.}
\label{tab:late-fusion}
\vskip -3pt
\renewcommand{\arraystretch}{1.}
\begin{adjustbox}{width=.8\linewidth}
\begin{tabular}{ccccccc}
\toprule
 & \textbf{Depth pre-fill} & \textbf{\# Points} & \textbf{Bad1 (\%)} & \textbf{Bad2 (\%)} & \textbf{Avg. Err. (px)} \\ \midrule
\ding{192} & -      & 0      & 5.01 & 0.41 & 0.374 \\
\ding{193} & -      & 300    & 4.32 & 0.37 & 0.359 \\ \midrule
\ding{194} & \textbf{IP-Basic~\cite{ku2018ipfill}} & \textbf{Dense}  & \textbf{3.61} & \textbf{0.36} & \textbf{0.310} \\ \midrule
\ding{195} & Neural net   & Dense  & 3.67  & \textbf{0.36} & 0.317 \\ \bottomrule 
\end{tabular}
\end{adjustbox}
% \vskip -20pt
\end{table}

\mypara{What pre-filling method should we apply?} Moving beyond the toy example to more realistic LiDAR guidance, a more sophisticated pre-filling method may be required. This problem is closely related to depth completion, where the goal is to predict a dense depth map from a sparse one, often using a monocular image as additional input. We explore two depth-completion methods: 
\begin{itemize}[nosep, topsep=2pt, parsep=2pt, partopsep=2pt, leftmargin=*]
    \item an image-processing-based interpolation method \cite{ku2018ipfill} (\ie, IP-Basic), without the need to train a neural network;
    \item a more accurate approach based on a neural network (3.26M parameters), inspired by~\cite{park2020non} (\cref{fig:unet}).\footnote{For training the neural network-based depth pre-filling model in our study, we follow the objective function and hyperparameters used in training GRAFT-Stereo, with the exception that the ground-truth signal is the depth map instead of the disparity map. The pre-filling network is pretrained prior to GRAFT-Stereo using 300 LiDAR points and is kept frozen across all experiments, including those with varying LiDAR sparsity conditions.}
\end{itemize} 
We fix the number of LiDAR points to 300 per frame, pre-train the neural network, and freeze it throughout the entire experiment. \cref{fig:overall} shows examples of both methods.

\cref{tab:prefill} reports the disparity error of the pre-filling methods alone, while \cref{tab:late-fusion} presents the results of LiDAR-guided RAFT-Stereo, using the pre-filled initial disparity maps. We highlight two key observations.
First, we observe that pre-filling the disparity map significantly boosts final performance (\ding{193}~\textit{vs.}~\ding{194} \& \ding{195}), demonstrating its effectiveness for LiDAR-guided RAFT-Stereo. 
Second, a more accurate depth completion method does not necessarily yield better performance (\ding{194}~\textit{vs.}~\ding{195}). 
This suggests that even a less accurate pre-filling method can benefit RAFT-Stereo, likely because RAFT-Stereo's iterative refinement can correct and improve upon the initial estimate.

\begin{figure}[t]
  \centering
  \vskip 2pt
  \includegraphics[width=.7\linewidth]{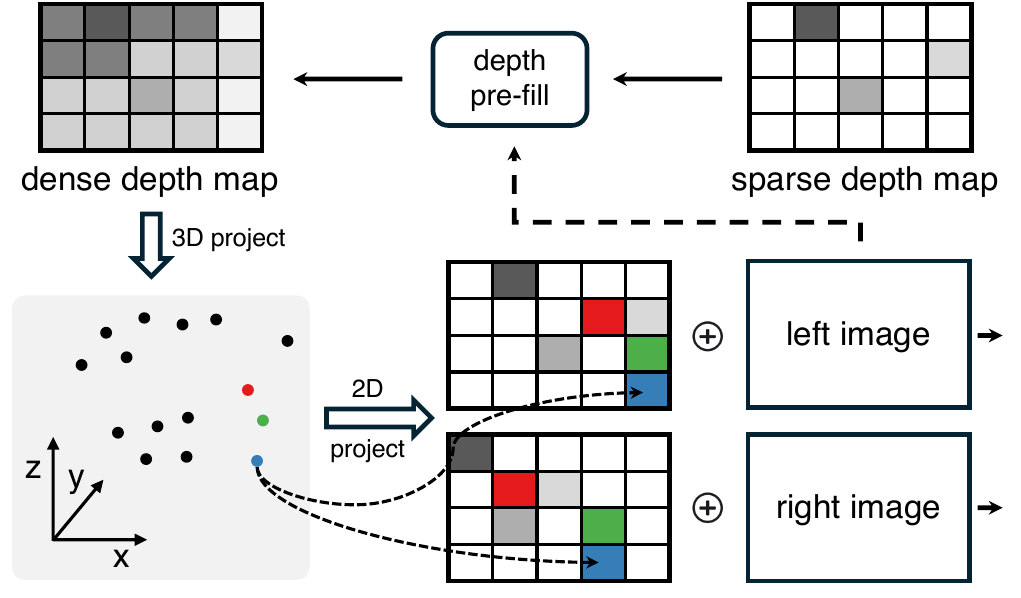}
  \caption{\textbf{Early fusion for enhanced cost volume.}
  Given the pre‑filled depth map, we back‑project pixels to 3D and
  re‑project them into both images. Each 3D point’s XYZ is concatenated
  with RGB from the left/right image. This fused tensor is passed to the feature encoder.}
  \label{fig:early-fusion}
  % \vspace{-10pt}
\end{figure}

\begin{table}[t]
  \centering
  \caption{\textbf{Incorporating LiDAR information via early fusion
  improves the LiDAR‑guided RAFT‑Stereo model introduced in
  \cref{ss:late}.} Interestingly, while depth pre‑filling yields
  additional gains, its underlying rationale differs fundamentally from
  that of densifying the initial disparity map—highlighting the need for
  a more accurate depth completion method. *: sampled from the pre‑filled
  depth map.}
  \label{tab:early-fusion}
  \vspace{-3pt}
  \renewcommand{\arraystretch}{1.}
  \begin{adjustbox}{width=.8\linewidth}
    \begin{tabular}{cccccc}
      \toprule
      & \textbf{Feat. encoder} & \textbf{Depth pre‑fill} & \textbf{\# Points} &
        \textbf{Bad1 (\%)} & \textbf{Avg. Err. (px)} \\ \midrule
      \ding{192} & Convolution    & –                     & 0        & 3.61 & 0.310 \\
      \ding{193} & Convolution    & –                     & 300      & 3.51 & 0.299 \\
      \ding{194} & Self‑attention & –                     & 0        & 3.56 & 0.304 \\
      \ding{195} & Self‑attention & –                     & 300      & 3.41 & 0.297 \\ \midrule
      \ding{196} & Self‑attention & IP‑Basic~\cite{ku2018ipfill} & Dense* & 9.90 & 0.522 \\ \midrule
      \ding{197} & \textbf{Self‑attention} & \textbf{Neural net} & \textbf{1,000}* & \textbf{3.33} & \textbf{0.294} \\
      \ding{198} & Self‑attention & Neural net            & Dense*   & 3.44 & 0.300 \\ \bottomrule
    \end{tabular}
  \end{adjustbox}
  \vspace{-10pt}
\end{table}

\subsection{Can we use sparse LiDAR guidance elsewhere?}
\label{ss:early}

In addition to improving cost volume retrieval, we investigate \emph{how to construct a more accurate cost volume} by incorporating sparse LiDAR guidance---ideally with minimal modification to the RAFT-Stereo pipeline. 

Since RAFT-Stereo builds the cost volume via feature correlation (\cref{eq:correlation}), pixels of the same 3D point should ideally correlate highly. We thus project each LiDAR point into both images and concatenate its 3D coordinates with the RGB values of the corresponding pixels. Assigning identical XYZ to these matched pixels injects geometric cues into the encoder, guiding it toward a more accurate cost volume.
\cref{fig:early-fusion} illustrates this idea. For pixels without a projected LiDAR point, we replace the missing 3D coordinates with zeros. We discuss depth pre-filling strategies later in this subsection.

\cref{tab:early-fusion} summarizes the experimental results. We begin with the LiDAR-guided RAFT-Stereo model introduced in \cref{ss:late}, which uses an initial disparity map obtained via image-processing-based depth pre-filling~\cite{ku2018ipfill}. First, we compare the default ResNet feature encoder using only image input against a variant that incorporates additional LiDAR information. As shown in \cref{tab:early-fusion}~\ding{192}~\textit{vs.}~\ding{193}, incorporating LiDAR points during feature encoding leads to a clear performance gain. Next, we investigate a Transformer-like feature encoder with self-attention~\cite{dosovitskiy2020image}, due to its ability to capture global context. 
This change yields a further improvement, as shown in \cref{tab:early-fusion}~\ding{192}~\textit{vs.}~\ding{194} and~\ding{193}~\textit{vs.}~\ding{195}.

\mypara{Is pre-filling helpful?}
We now investigate whether densifying the LiDAR guidance can further improve the early fusion approach described above. Following~\cref{ss:late}, we consider both image-processing-based (\ie, IP-Basic) and neural-network-based depth completion methods. In sharp contrast to the findings in~\cref{ss:late}, neither method improves stereo matching performance (\cref{tab:early-fusion} \ding{196} \& \ding{198}~\textit{vs.}~\ding{195}); in fact, IP-Basic pre-filling significantly increases disparity error. 

We hypothesize that the effectiveness of depth pre-filling differs between late fusion and early fusion. In the late fusion case, the goal of pre-filling the initial disparity map is to retrieve informative and smooth cost volume features for iterative disparity refinement. In contrast, for early fusion, LiDAR guidance is injected into the input image space to explicitly indicate pixel correspondences---requiring much higher accuracy. As a result, inaccurate or overly smoothed depth completion can degrade performance rather than help.

To address this issue, we subsample the depth predictions from the completion network, retaining only the top-confidence pixels based on the prediction confidence from the final block~\cite{park2020non}. We find that retaining the top 1,000 depth-completed points yields further improvements in our final LiDAR-guided RAFT-Stereo (\textbf{GRAFT-Stereo}) model, which integrates both early and late fusion strategies.

\section{Additional Experiments and Analysis}

\begin{table}[]
\centering
\caption{\textbf{Quantitative comparison to the SotA LiDAR-guided stereo methods on the KITTI Depth Completion dataset~\cite{uhrig2017kittidc}.} 
We train separate models, including the compared methods based on their official code for each guidance condition.
GRAFT-Stereo consistently outperforms prior methods across various sparse LiDAR settings in depth metrics. gd: naive LiDAR-guidance; $^\dagger$: uniform sampling; $^\S$: beam sampling.}
\label{tab:sota}
\begin{adjustbox}{width=\linewidth}
\begin{tabular}{lccc|ccc}
\toprule
 & \textbf{300 points}$^\dagger$ & \textbf{1000 points}$^\dagger$ & \textbf{3000 points}$^\dagger$ & \textbf{4-beam}$^\S$ & \textbf{8-beam}$^\S$ & \textbf{16-beam}$^\S$ \\ \cmidrule(lr){2-2}
\cmidrule(lr){3-3}
\cmidrule(lr){4-4}
\cmidrule(lr){5-5}
\cmidrule(lr){6-6}
\cmidrule(lr){7-7}
\textbf{Method} & \textbf{RMSE} / \textbf{MAE} & \textbf{RMSE} / \textbf{MAE} & \textbf{RMSE} / \textbf{MAE} & \textbf{RMSE} / \textbf{MAE} & \textbf{RMSE} / \textbf{MAE} & \textbf{RMSE} / \textbf{MAE} \\ \midrule
EG-Depth~\cite{xu2023egdepth} & 829.7 / 283.6 & 822.6 / 263.0 & 764.0 / 242.5 & 900.0 / 307.3 & 851.2 / 277.8 & 858.9 / 276.7 \\
SDG-Depth~\cite{li2024sdgdepth} & 883.9 / 307.2 & 816.7 / 290.7 & 763.8 / 259.2 & 876.8 / 326.9 & 789.2 / 279.5 & 792.6 / 279.7 \\
RAFT-Stereo~\cite{lipson2021raftstereo}~+~gd & 929.5 / 297.0 & 835.0 / 274.8 & 827.5 / 243.1 & 863.8 / 297.4 & 874.7 / 293.6 & 873.3 / 288.5 \\
GRAFT-Stereo (late) & 817.7 / 237.5 & 747.0 / 210.1 & 717.8 / 193.9 & 817.9 / 255.1 & 803.0 / 243.2 & 808.0 / 241.1 \\
GRAFT-Stereo (full) & \textbf{739.5} / \textbf{220.2} & \textbf{716.5} / \textbf{208.3} & \textbf{675.6} / \textbf{186.2} & \textbf{779.0} / \textbf{240.2} & \textbf{774.4} / \textbf{235.1} & \textbf{767.8} / \textbf{230.5} \\ \bottomrule
\end{tabular}
\end{adjustbox}
\vskip -5pt
\end{table}

\begin{figure}[t]
    \centering
    \includegraphics[width=\linewidth]{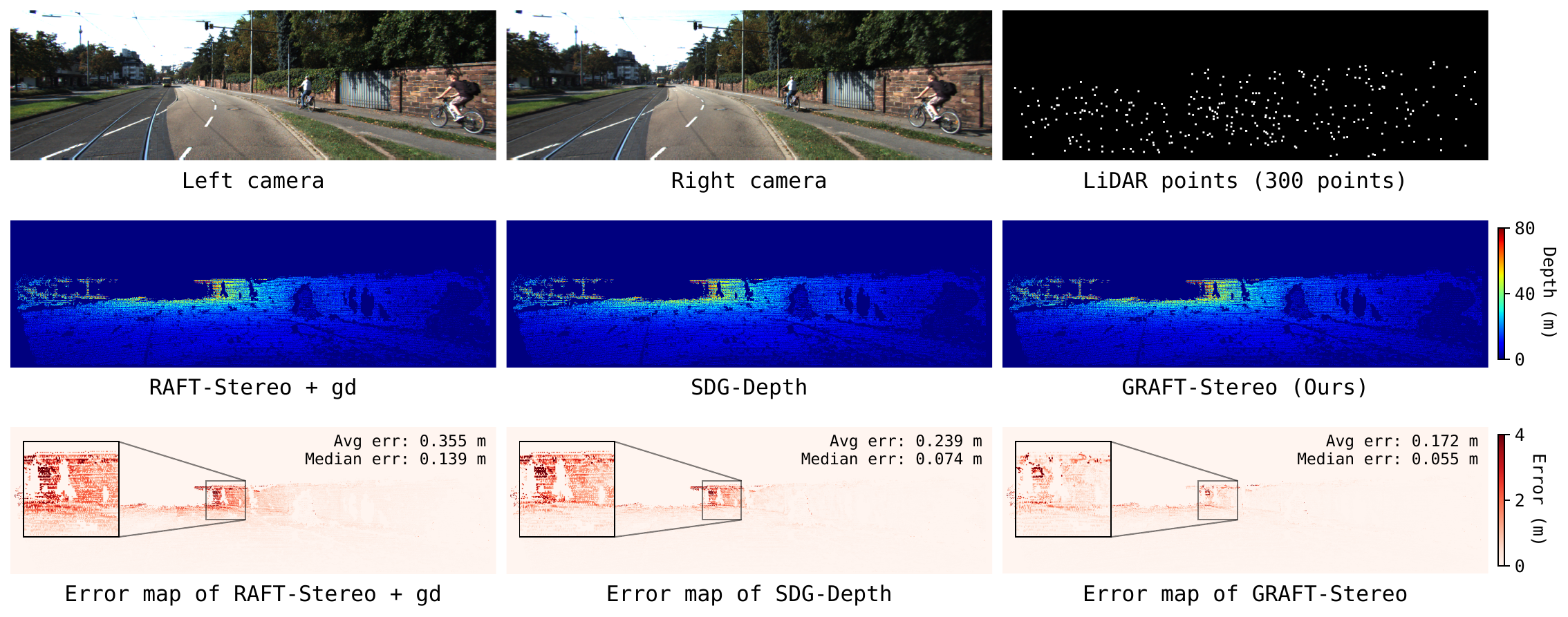}
    \caption{\small\textbf{Qualitative comparison of predicted depth maps.} GRAFT-Stereo produces more accurate depth predictions. Error maps are computed with respect to ground-truth. gd: naive LiDAR-guidance.}
\label{fig:quali-depth}
\end{figure}

\begin{table}[t]
\centering
\caption{\textbf{Eveluation on different depth ranges.} GRAFT-Stereo leads to improved depth accuracy across both near and distant ranges.} 
\label{tab:exp_depth_range}
\renewcommand{\arraystretch}{1}
\begin{adjustbox}{width=\linewidth}
\begin{tabular}{ccccc}
\toprule
 & \textbf{RAFT-Stereo~\cite{lipson2021raftstereo}} & \textbf{RAFT-Stereo~\cite{lipson2021raftstereo} + gd} & \textbf{GRAFT-Stereo (late)} & \textbf{GRAFT-Stereo (full)} \\
\cmidrule(lr){2-2}
\cmidrule(lr){3-3}
\cmidrule(lr){4-4}
\cmidrule(lr){5-5}
\textbf{Depth range} & \textbf{MAE / RMSE} & \textbf{MAE / RMSE} & \textbf{MAE / RMSE} & \textbf{MAE / RMSE} \\ \midrule
0 -- 30m & 160.5 / 400.8 & 155.6 / 384.1 & 129.9 / 396.3 & \textbf{122.6} / \textbf{339.5} \\
30 -- 100m & 1545.2 / 2525.3 & 1514.5 / 2770.7 & 1189.7 / 2277.0 & \textbf{1082.3} / \textbf{2121.1} \\ \bottomrule
\end{tabular}
\end{adjustbox}
\end{table}

\begin{table}[t]
% \vskip -15pt
% \begin{minipage}{.42\textwidth}
\centering
\caption{\textbf{Depth estimation results on VKITTI2~\cite{cabon2020vkitti} and MS2 \cite{shin2023ms2}}. \#~Points: available LiDAR points per frame. GRAFT-Stereo surpasses other methods, demonstrating its generalizability in leveraging sparse LiDAR points. We report results using the full fusion version of GRAFT-Stereo.} 
\label{tab:exp_vkitti_ms2}
\renewcommand{\arraystretch}{1}
\begin{adjustbox}{width=.7\linewidth}
\begin{tabular}{llccccc}
\toprule
\textbf{Dataset} & \textbf{Method} & \textbf{\# Points} & \textbf{RMSE (mm)} & \textbf{MAE (mm)}  \\ \midrule
\multirow{3}{*}{VKITTI2} & EG-Depth~\cite{xu2023egdepth} & 300 & 4437.8  & 927.5  \\
 & SDG-Depth~\cite{li2024sdgdepth} & 300 & 3818.7 & 1187.7 \\
 & GRAFT-Stereo & 300 & \textbf{3761.7} & \textbf{877.5} \\ \midrule 
\multirow{3}{*}{MS2} & EG-Depth~\cite{xu2023egdepth} & 300 & 1854.5  & 998.5  \\
 & SDG-Depth~\cite{li2024sdgdepth} & 300 & 1471.1  & 843.3  \\
 & GRAFT-Stereo & 300 & \textbf{1465.9}  & \textbf{781.4}   \\ \bottomrule 
\end{tabular}
\end{adjustbox}
% \end{minipage}
\vskip -5pt
\end{table}

\begin{table}[t]
\centering
% \caption{\textbf{Comparison with state-of-the-art LiDAR-guided stereo methods on the KITTI Depth Completion dataset~\cite{uhrig2017kittidc} under denser 64-beam LiDAR setting.} GRAFT-Stereo still demonstrates competitive performance under dense LiDAR setting. gd: naive guidance.}
\caption{\textbf{Comparison on the KITTI Depth Completion dataset~\cite{uhrig2017kittidc} under the 64-beam LiDAR setting.} 
GRAFT-Stereo remains competitive even with dense LiDAR input. 
gd: naive guidance.}

\label{tab:suppl-64-beam}
\vskip-3pt
\renewcommand{\arraystretch}{1}
\begin{adjustbox}{width=.8\linewidth}
\begin{tabular}{lcccc}
\toprule
\textbf{Method} & \textbf{RMSE (mm)} & \textbf{MAE (mm)} & \textbf{Bad1 (\%)} & \textbf{Avg. Err (px)} \\ \toprule
EG-Depth~\cite{xu2023egdepth}   & 675.5   & 197.2 & 3.68 & 0.302   \\
SDG-Depth~\cite{li2024sdgdepth}  & \textbf{623.2}  & 197.6 & \textbf{3.60} & 0.300 \\
RAFT-Stereo~\cite{lipson2021raftstereo} + gd  & 768.7 & 228.6 & 4.12 & 0.317 \\ 
GRAFT-Stereo (late fusion) & 697.4 & \textbf{189.9} & 3.80 & \textbf{0.291} \\
GRAFT-Stereo (full fusion) & 685.4 & 193.0 & 4.12 & 0.306 \\
\bottomrule
\end{tabular}
\end{adjustbox}
\vskip-5pt
\end{table}

\begin{figure}[t]
\centering
\includegraphics[width=\linewidth]{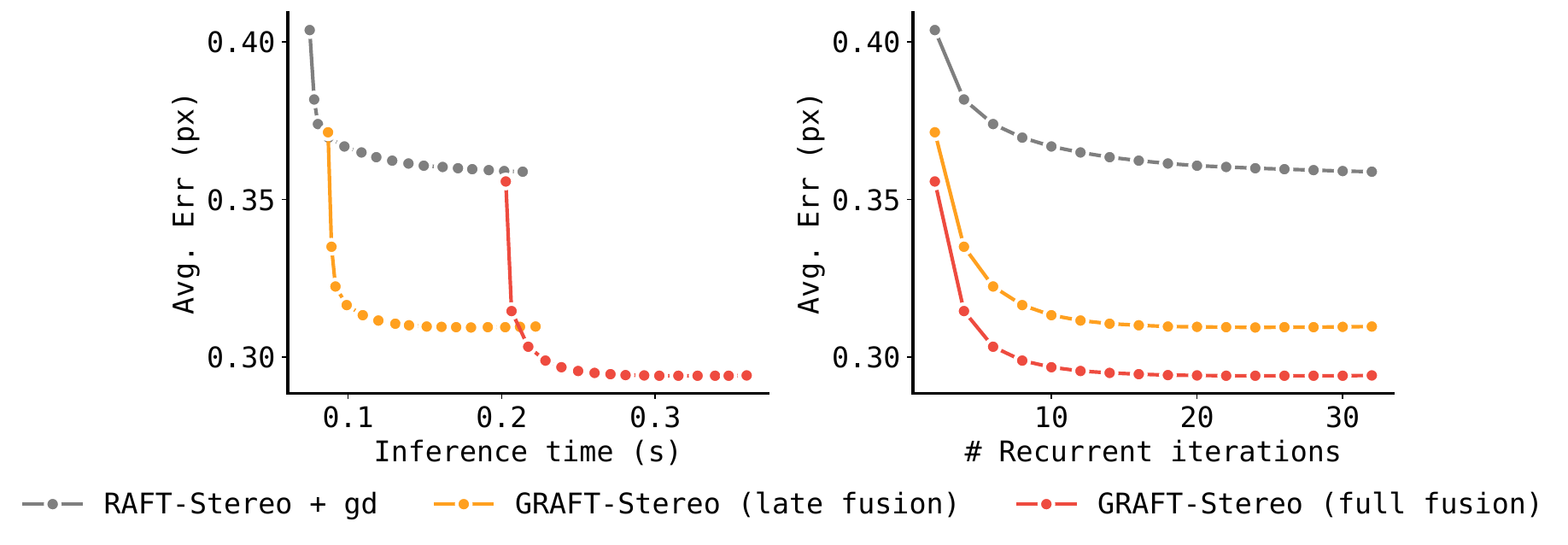}
\vskip-5pt
\caption{\textbf{Computation and accuracy trade-off during test time.} 
We compare accuracy and inference time by varying the number of recurrent iterations. GRAFT-Stereo variants provide a broad range of speed–accuracy trade-offs.}
\label{fig:suppl_computation_acc}
\end{figure}

\mypara{Comparison to SotA LiDAR-guided methods.}
Beyond the disparity results in~\cref{fig:1b}, we evaluate GRAFT-Stereo on depth estimation against representative LiDAR-guided stereo methods~\cite{xu2023egdepth,li2024sdgdepth}. 
We retrain all baselines using their official implementations, with separate models for each sparsity level. 
Following~\cref{tab:early-fusion}, we also sample 1{,}000 points from the pre-filled depth map for full fusion.
As shown in~\cref{tab:sota}, our method consistently outperforms prior approaches under sparse LiDAR input. 
A qualitative comparison in~\cref{fig:quali-depth} also shows that GRAFT-Stereo produces more accurate depth maps, especially at longer ranges.

\mypara{Evaluation across depth ranges.}
To assess performance across different depths, we evaluate GRAFT-Stereo using the 300-point sparse LiDAR setting. As shown in~\cref{tab:exp_depth_range}, GRAFT-Stereo consistently improves upon RAFT-Stereo in both near and far ranges, outperforming the naive use of sparse LiDAR.

\mypara{Evaluation on additional datasets.}
We further evaluate our method on additional datasets, including VKITTI2~\cite{cabon2020vkitti} and MS2~\cite{shin2023ms2}. 
We fine-tune all models on each dataset using their official implementations. 
As shown in~\cref{tab:exp_vkitti_ms2}, GRAFT-Stereo consistently outperforms the baselines across both datasets, demonstrating the strength of RAFT-Stereo and the effectiveness of our sparse-guidance strategy.

\mypara{Analysis of recurrent iterations and runtime.}
In~\cref{fig:suppl_computation_acc}, we illustrate the trade-off between accuracy and inference time by varying the number of recurrent iterations. 
We include the CPU runtime of IP-Basic~\cite{ku2018ipfill} for image-processing-based pre-filling. 
Importantly, our main late-fusion variant remains fully convolutional and image-based, and therefore introduces negligible additional computational overhead. 
While all models benefit from additional iterations, GRAFT-Stereo variants provide a broad speed–accuracy trade-off, with the full model achieving the highest accuracy.

\mypara{Comparison under denser LiDAR settings.}
While our focus is \emph{extremely sparse} LiDAR, we also report 64-beam results for completeness. 
As shown in~\cref{tab:suppl-64-beam}, GRAFT-Stereo remains competitive with prior SotA methods. 
Interestingly, late fusion outperforms full fusion, as our early-fusion design can oversmooth when dense LiDAR provides similar depths across neighboring pixels.

\section{Conclusion and Discussion}
\label{sec:conclusion}
We study LiDAR-guided RAFT-Stereo under extreme sparsity.
We show that its cost-volume retrieval in late fusion can suppress sparse LiDAR cues, and that pre-filling the initial disparity map effectively mitigates this issue.
This simple strategy yields substantial gains in the extremely sparse regime, while early fusion further improves performance when needed.
Together, these components form \textbf{GRAFT-Stereo}, which achieves state-of-the-art results across multiple datasets under sparse LiDAR conditions.
Future work includes generalizing our analysis to foundation stereo models~\cite{wen2025foundationstereo}, analyzing how the spatial distribution of LiDAR points affects guidance, and improving temporal consistency.

\section*{Acknowledgment}
    This research is supported in part by a grant from the Office of Naval Research DoD (N68335-23-C-0704).
    We thank the Ohio Supercomputer Center for its generous support in providing computational resources.
	
%%%%%%%%%%%%%%%%%%%%%%%%%%%%%%%%%%%%%%%%%%%%%%%%%%%%%%%%%%%%%%%%%%
%\addtolength{\textheight}{-12cm}
%\vspace{10mm}
\bibliographystyle{IEEEtran}
% Your .bib file here
\bibliography{root} 

\end{document}